\documentclass[10pt,twocolumn,letterpaper]{article}

\usepackage{iccv}
\usepackage{times}
\usepackage{epsfig}
\usepackage{graphicx}
\usepackage{amsmath}
\usepackage{amssymb}
\usepackage{multirow}
\usepackage{stfloats}
\usepackage{tabularx}
\usepackage{multirow}
\usepackage{booktabs}
\usepackage[absolute,overlay]{textpos}

\usepackage[bookmarks=false,breaklinks=true]{hyperref}
\iccvfinalcopy 

\def\iccvPaperID{****} 

\begin{document}
\title{Annotation-Free Group Robustness via Loss-Based Resampling}

\author{Mahdi Ghaznavi\\
{\tt\small mahdi.ghaznavi@ce.sharif.edu}
\and
Hesam Asadollahzadeh\\
{\tt\small hesam.asadzadeh26@researcher.sharif.edu}
\and
HamidReza Yaghoubi Araghi\\
{\tt\small hamidreza.yaghoubiaraghi@sharif.edu}
\and
Fahimeh HosseiniNoohdani\\
{\tt\small fhosseini@ce.sharif.edu}
\and
Mohammad Hossein Rohban\\
{\tt\small rohban@sharif.edu}
\and
Mahdieh Soleymani Baghshah\\
{\tt\small soleymani@sharif.edu}
\and
Sharif University of Technology\\
Tehran, Iran
}

\maketitle
\ificcvfinal\thispagestyle{plain}\fi

\begin{abstract}
It is well-known that training neural networks for image classification with empirical risk minimization (ERM) makes them vulnerable to relying on spurious attributes instead of causal ones for prediction. Previously, deep feature re-weighting (DFR) has proposed retraining the last layer of a pre-trained network on balanced data concerning spurious attributes, making it robust to spurious correlation. However, spurious attribute annotations are not always available. In order to provide group robustness without such annotations, we propose a new method, called loss-based feature re-weighting (LFR), in which we infer a grouping of the data by evaluating an ERM-pre-trained model on a small left-out split of the training data. Then, a balanced number of samples is chosen by selecting high-loss samples from misclassified data points and low-loss samples from correctly-classified ones. Finally, we retrain the last layer on the selected balanced groups to make the model robust to spurious correlation.
For a complete assessment, we evaluate LFR on various versions of Waterbirds and CelebA datasets with different spurious correlations, which is a novel technique for observing the model's performance in a wide range of spuriosity rates. While LFR is extremely fast and straightforward, it outperforms the previous methods that do not assume group label availability, as well as the DFR with group annotations provided, in cases of high spurious correlation in the training data.
\end{abstract}
\\
\section{Introduction}

Neural networks have the tendency to rely on patterns in input data that are highly correlated with the target class in image classification. If the correlations that a model relies on are spurious, i.e. non-causal, the model's performance will drop under spurious correlation shift in the test time~\cite{xiao2021noisew,Li_2023_CVPR_Whac_A_Mole,singla2022salient}. In such situations, the model performs well on \textit{majority groups} of training data which have the spurious pattern, while the performance is poor on \textit{minority groups}~\cite{pmlr-v81-buolamwini18a,doi:10.1089/big.2016.0047}. As an example, in Waterbirds~\cite{Sagawa2019DistributionallyRN} dataset,  there is a spurious correlation between the waterbird/landbird label and the water/land background, which implies waterbirds on water and landbirds on land as the majority groups and the other two combinations as minority groups. The final goal is to achieve the best \textit{worst group accuracy (WGA)} among all groups for the sake of robustness to correlation shift.

Some earlier methods consider accessing group annotations for the whole data~\cite{Sagawa2019DistributionallyRN,Kirichenko2022LastLR} or a small portion of it~\cite{sohoni2022barack,lee2023diversify}. ~\cite{Sagawa2019DistributionallyRN} proposes group distributionally robust optimization (GDRO), which softly optimizes the worst-case loss among groups during training.
~\cite{Kirichenko2022LastLR} suggests that retraining the last layer of a classification model on a balanced dataset concerning group annotation and label is sufficient for making the model robust to spurious correlation shift. Their deep feature re-weighting (DFR) method has achieved good performance while being simple and fast. Nonetheless, group annotations are not always available.
Some works do not assume such annotation requirement ~\cite{pmlr-v139-liu21f,nam2020learning,Sohoni2020NoSL,Creager2020EnvironmentIF,Liu2021HeterogeneousRM}, but they need to train or fine-tune the whole model to achieve robustness.


As stated in ~\cite{Kirichenko2022LastLR}, models trained with empirical risk minimization (ERM) learn core (causal) features besides the spurious ones, regardless of how strong the spurious correlation is. 
Thus, it seems that all we need is to find the proper linear combination of features to achieve the desired WGA. 
Recently, ~\cite{osti_10437778} proposed automatic feature re-weighting (AFR), which re-weights each sample while training the model's last layer, based on the model's probability of predicting the correct label on a given sample. However, as their weighting does not consider the size of minority groups, it does not work in situations where the size of the minority group is too small. Our work is focused on addressing this issue. 

\begin{figure*}[t]
\begin{center}
\includegraphics[width=\linewidth]{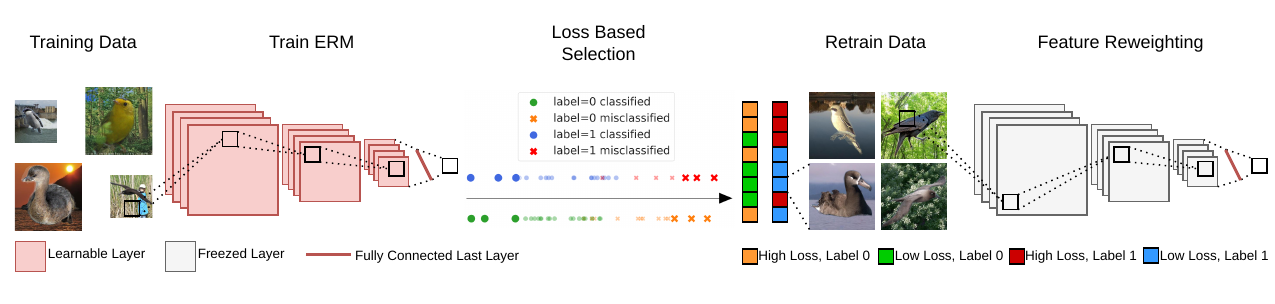}
\end{center}
   \caption{Overview of LFR method. After training a model on unbalanced data, we evaluate it on a small portion of the dataset (about 20\% of the total). Then, we select images with low loss values in correctly-classified images and high loss in misclassified images and select an equal number of them in each class. At last, retraining the last layer of the model on the balanced dataset improves the model's robustness.}
\label{fig:abstract}
\end{figure*}

In this work, we propose a new method for robustness to spurious correlation shifts. First, we infer groups based on the model's correctness of prediction on a set of data. We then select a few number of high-loss samples from the misclassified ones and low-loss samples from the correctly-classified images in each group. 
Finally, we apply DFR for feature re-weighting based on the selected groups. Our method, called loss-based feature re-weighting (LFR) is a simple yet effective method that achieves state-of-the-art (SOTA) on various famous datasets with high spurious correlation. The overview of LFR is shown in Figure \ref{fig:abstract}. 

Using LFR at the end of any training pipeline can improve WGA free of charge since it is fast, straightforward, and doesn't assume any inductive bias. LFR works well on a wide range of high to low spurious correlations by using only a small subset of training data for last layer retraining.
By visualizing the saliency map of the model before and after last-layer retraining, we observe that the model's attention regions shift from spurious to core attributes in both majority and minority samples.


\section{Related Work}
We can divide the related work in the field by different means.

\textbf{Availability of group annotations}: Some works in the literature have an assumption of access to group annotations~\cite{Sagawa2019DistributionallyRN,Kirichenko2022LastLR}. On the other hand, some works have tried to make a model robust to group shift without the availability of group annotations~\cite{pmlr-v139-liu21f,nam2020learning,pmlr-v162-zhang22z,osti_10437778}. 

There are also some works that only need a small amount of group-annotated data.~\cite{sohoni2022barack} used these group annotations to train a classifier for predicting group labels of the remaining data points, which are then fed into GDRO.
~\cite{lee2023diversify} utilizes a set of diverse classifiers on a dataset without using group annotations, and then annotates a few samples for selecting the favorable model among this set. 

\textbf{Scope of Training}:
Different methods have different scopes of training, which leads to different amounts of required computation and effectiveness. Some works make a model robust during initial training. GDRO~\cite{Sagawa2019DistributionallyRN} solves a two-level optimization to minimize the maximum loss among all groups.~\cite{nam2020learning} proposes learning from failure (LfF), which trains two models: a super-biased model to majority groups, and another model that upweights images with high loss value in a biased model. Thus, the second model goes in a way to be robust.
\\
On the other hand, some works fine-tune a biased pre-trained model that had been trained by ERM.~\cite{pmlr-v139-liu21f} proposes just training twice (JTT), which upweights misclassified data points during the retraining.~\cite{pmlr-v162-zhang22z} uses a contrastive method to make samples of a class that are correctly or misclassified by a pre-trained model close together. Both aforementioned methods need validation data for tuning weighting hyperparameters.
\\
However, some methods just retrain only the last layer of  pre-trained models. These methods are based on DFR~\cite{Kirichenko2022LastLR}, which suggest retraining the last layer of the model on group-balanced data to make the model robust to spurious correlation. Other methods were proposed to solve the requirement of group annotation by different techniques for grouping or weighting.
~\cite{labonte2022dropout} suggests putting samples of each class whose prediction changed after dropout in one group and the rest in another for using DFR. AFR~\cite{osti_10437778} is a recently proposed method in this line, which upweights samples with lower confidence on the ground truth label during retraining the last layer.
\\
Our LFR method does not need access to group annotations and retrains only the last layer of a model which is pre-trained with ERM. Our setting is similar to AFR, and we compare it to LFR.\\

\section{Method}

\subsection{Setting}
Consider an image classification task among two classes on a dataset $\mathcal{D}\subset{\mathbb{R}^{3\times H\times W}}$ where $H$ and $W$ are the height and width of input images. Suppose images of each class $i\in\{1, 2\}$ in the dataset are split into unknown majority $Maj_i$ and $Min_i$ groups such that $|Maj_i|\gg|Min_i|$ and
images in $Maj_i$ have predictive attributes which are simpler than the core attribute of class $i$. We first train a deep neural network $f:\mathcal{\mathbb{R}}^{3\times H\times W}\rightarrow{}[0, 1]^2$ with ERM and loss function $\mathcal{L}:[0, 1]^2\rightarrow{}\mathbb{R}$ on a randomly selected 80\% of $\mathcal{D}$ with class balancing. Class balancing is a conventional necessary technique for training a good model in any classification task, even without strong spurious correlation. The function $f=softmax(h\circ g)$ is composed of a feature extractor $g:\mathbb{R}^{3\times H\times W}\rightarrow{}\mathbb{R}^d$ and a linear last layer function $h:\mathbb{R}^d\rightarrow{}\mathbb{R}^2$ with a softmax function on top of them. Then we are going to retrain the last layer $h$ of the model on the remaining 20\% of $\mathcal{D}$, which refers to it by $\mathcal{D}_{ll}$. We hope to have a higher WGA in groups $\{Maj_1, Maj_2\}\cup\{Min_1, Min_2\}$ than before retraining.

\subsection{Loss-Based Feature Reweighting}
As it is illustrated in Fig. \ref{fig:abstract}, after initial training, we evaluate the model $f$ on $\mathcal{D}_{ll}$. We refer to the correctly-classified and misclassified samples of class $i$ by $C_i$ and $M_i$. As we explain later in this section, these sets are good representatives of $Maj_i$ and $Min_i$ groups respectively.

After that, as has been suggested by~\cite{Kirichenko2022LastLR}, we want to select an equal number $s$ of data from each group in the set $\mathcal{G}=\bigcup_{i=1}^2 \{M_i, C_i\}$. $s$, the number of groups' samples, is a hyperparameter and can be any number, but we usually select it approximately as $s=min(|M_1|,|M_2|)$. The optimal amount $s$ is found by hyperparameter selection based on worst group accuracy on the validation set.
In each majority group like $C_i$, we select $s$ samples with the highest loss value $\mathcal{L}(f(x))$, whereas, for a minority group $M_i$, we select $s$ images with the lowest value of loss function.
Then, as has been suggested by~\cite{Kirichenko2022LastLR}, we reinitialize the last layer $h$ of the model and then train it on a balanced subset of $\mathcal{D}_{ll}$ which consists $s$ samples from each group in $\mathcal{G}$.

\begin{figure}
\begin{center}
\fbox{
\includegraphics[width=0.9\linewidth]{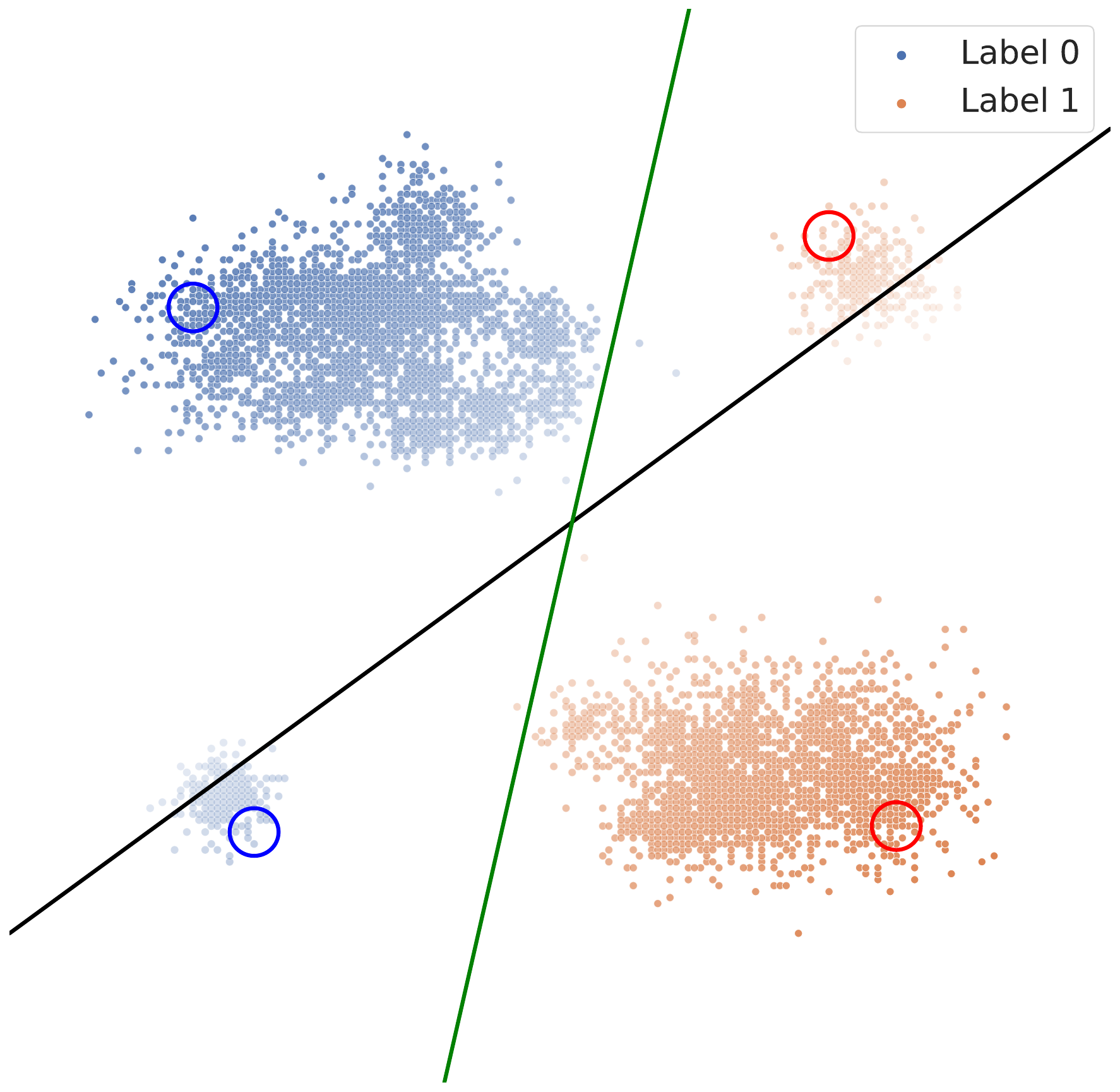}
}
\end{center}
   \caption{An abstraction of our LFR method in 2D space. The majority groups of data make the initial classifier biased towards themselves (black line). Selecting an equal number of samples from high-loss value misclassified and low-loss classified samples mitigates the bias towards a more robust classifier (green line)}
\label{fig:classifier}
\end{figure}

It is important to select samples based on loss value, not
randomly. If a model is pre-trained with ERM on a biased
dataset, it probably relies on spurious correlation for pre-
diction. Thus, the loss value of the model is a good sign of
belonging to majority or minority groups. So, in classified
samples, if a sample has low loss, there is a great chance
that the sample has a spurious attribute which has made the
model biased. On the other hand, in a misclassified group, if
a sample has a very high loss, it probably belongs to the
minority group without the spurious attribute. The success of
loss-based selection in comparison to random selection in
the ablation study also suggests that among correctly-classified samples, the ones that rely on spurious attributes have lower loss value than the ones that attend to core attributes. It is true as well for misclassified samples, as those with higher loss values are more likely to have not the spurious attribute. 
Figure \ref{fig:classifier} shows an abstraction of LFR in 2D space, which is compatible with the above explanations. Practical experiments also confirm our descriptions.

\section{Experiments}

\begin{table*}[htb]
\centering
\caption{Evaluation of worst and mean (in parenthesis) group accuracy on the Waterbirds dataset with various amounts of spurious correlation.}
\begin{tabular}{lcccccccl} 
 \toprule
  Method & 70\% & 80\% & 90\% & 95\% & 99\% & 100\%\\ [0.5ex] 
  \midrule
  ERM & 84.42 (93.13) & 83.42 (90.97) & 76.36 (88.06) & 71.31 (85.57) & 48.74 (75.73) & 30.33 (66.83) \\
 DFR & 91.59 (94.29) & 87.32 (91.73) & 88.43 (91.16) & 88.79 (92.48) & 83.80 (91.30) & 83.33 (91.20) \\
 \midrule
 AFR & \textbf{86.71} & 83.43 & 76.42 & 77.40 & 78.82 & 84.73 \\
 \midrule
 CFR (Ours) & 75.45 (87.15) & 78.62 (87.57) & 75.09 (84.93) & 67.45 (80.12) & 78.76 (87.08) & 82.09 (89.48) \\
 LFR (Ours) & 79.67 (85.25) & \textbf{83.81} (87.88) & \textbf{86.45} (91.02) & \textbf{86.85} (91.25) & \textbf{85.31} (89.24) & \textbf{86.29} (90.54) \\
 \bottomrule
\end{tabular}
\label{table:waterbirds}
\end{table*}

\begin{table*}[htb]
\centering
\caption{Evaluation of worst and mean (in parenthesis) group accuracy on the CelebA dataset with various amounts of spurious correlation.}
\begin{tabular}{lcccccccl} 
  \toprule
  Method & 70\% & 80\% & 90\% & 95\% & 99\% & 100\%\\ [0.5ex] 
  \midrule
  ERM & 62.22 (93.56) & 62.22 (92.81) & 61.11 (92.65) & 57.22 (93.28) & 45.56 (93.33) & 40.00 (93.16) \\
 DFR & 74.45 (92.71) & 77.44 (92.53) & 77.22 (91.71) & 76.11 (91.76) & 80.56 (91.42) & 74.44 (91.29) \\
 \midrule
 AFR & 84.06\% & 85.34\% & 78.89\% & 66.30\% & 82.41\% & 77.96\% \\
 \midrule
 CFR (Ours) & \textbf{86.67} (88.26) & 85.30 (85.17) & 80.00 (84.40) & 63.89 (86.97) & 83.59 (85.17) & 82.56 (84.83)\\
 LFR (Ours) & 83.89 (87.20) & \textbf{86.73} (88.06) & \textbf{85.35} (84.26) & \textbf{79.36} (81.09) & \textbf{84.63} (86.06) & \textbf{82.78} (84.93) \\
 \bottomrule
\end{tabular}
\label{table:celeba}
\end{table*}

\subsection{Dataset}
We evaluate LFR on two popular datasets.

\textbf{Waterbirds}: This dataset has been introduced by~\cite{Sagawa2019DistributionallyRN}. It is a classification between waterbirds and landbirds, where most of the waterbirds are in the sea and most of the landbirds are seen in the land background. Here, the background scene is considered a spurious feature in both classes.

\textbf{CelebA}: Using the dataset which has been proposed by~\cite{7410782}, CelebA is a hair color classification of celebrities dataset. In this dataset, most celebrities with blonde hair are women. So, only the blond class has a spurious correlation between sex and the color of hair. CelebA has also some label noise, i.e., some labels are not correct as we have seen. 

Unlike previous methods which only have considered standard versions of these datasets, we have created six versions of these datasets with various spurious correlations. All datasets have four partitions for initial training (\textit{Tr}), last layer retraining (\textit{LLR}), validation (\textit{V}), and test (\textit{Te}). In each dataset, the size of the \textit{LLR} partition is approximately $1/4$ of the size of \textit{Tr} data. Also, the spurious correlation is equal in \textit{LLR} and \textit{Tr} data, except for datasets with 99\%, and 100\% spurious correlation, in which the \textit{LLR} has 95\% spurious correlation. This is because we need a few samples of minority groups in \textit{LLR} data, which can not be satisfied when less than 1\% of data belongs to them unless we have much more data than we already have. 

\subsection{Setup}


 Similar to previous works~\cite{Sagawa2019DistributionallyRN}~\cite{Kirichenko2022LastLR}~\cite{pmlr-v139-liu21f}, we have used ResNet-50~\cite{7780459} architecture which had been pre-trained on ImageNet~\cite{DenDon09Imagenet}. We have changed the last linear layer of the model to match our output dimension. 
Then, we trained the model on the \textit{Tr} partition of each dataset until convergence had been achieved. After reinitializing the last linear layer of the model, we have used the setting in~\cite{Kirichenko2022LastLR} for retraining the last layer. We have used validation data \textit{V} for hyperparameter tuning. Finally, we have reported the resulting WGA on test partition \textit{Te}. More details about experiments' settings are proposed in \ref{setup_det}.

For AFR, we ran their source code and selected the best hyperparameter $\gamma$, which specifies how much to upweights examples with poor predictions. It gives an extra and maybe unfair advantage to AFR, as larger $\gamma$ is suitable for high spurious correlation, and lower $\gamma$ is better for low one. So, selecting the best $\gamma$ enables extra information about the amount of spurious correlation. Nevertheless, we outperform AFR with this setting except for low spurious correlation.

\subsection{Results}

\begin{table*}[htb]
\caption{Effect of loss-based selection and random selection in each spurious correlation of Waterbirds dataset on worst and mean (in parentheses) group accuracy. Results are in percentage. CR-ML and CL-MR stand for \textit{\textbf{C}orrectly classified with \textbf{R}andom selection- \textbf{M}isclassified with \textbf{L}oss-based selection} and
\textit{\textbf{M}isclassified with \textbf{L}oss-based selection-\textbf{C}orrectly classified with \textbf{R}andom selection} respectively. While when the spurious correlation is not lower than 90\% LFR has the best performance, when the correlation is 70\% or 60\% random selection of misclassified samples leads to a higher WGA.}
\centering
\begin{tabular}{lcccccl} 
 \toprule
 Method & 70\% & 80\% & 90\% & 95\% & 99\% & 100\%\\ [0.5ex] 
 \midrule
 LFR & 79.67 (85.25) & 83.81 (87.88) & \textbf{86.45} (91.02) & \textbf{86.85} (91.25) & \textbf{85.31} (89.24) & \textbf{86.29} (90.54) \\
 CFR  & 75.45 (87.15) & 78.62 (87.57) & 75.09 (84.93) & 67.45 (80.12) & 78.76 (87.08) & 82.09 (89.48) \\
 LFR-CR-ML & 81.09 (89.61) & 83.36 (88.65) & 82.68 (88.44) & 83.80 (90.39) & 80.84 (87.93) & 83.80 (90.07) \\
 LFR-CL-MR & \textbf{84.65} (90.27) & \textbf{84.27} (91.21) & 80.37 (88.89) & 86.45 (92.03) & 83.96 (88.87) & 83.96 (89.35) \\
 \bottomrule
\end{tabular}
\label{table:ablation}
\end{table*}

The results of Waterbirds and CelebA datasets 
are shown in tables \ref{table:waterbirds} and \ref{table:celeba} respectively.
As can be observed, when the spurious correlation is high enough, LFR outperforms the previous AFR method and even the DFR baseline, in which we have access to the ground truth for the groups. 

We also put the method to the test by randomly selecting an equivalent number of samples from both the correctly classified and misclassified data in each class. This simplified version of our technique is termed Classification-based Feature Re-weighting (CFR). As illustrated in table \ref{table:ablation}, the loss-based selection is important in both correctly classified and misclassified groups. However, when the spurious correlation is low, the impact of loss-based selection diminishes, particularly for misclassified samples. It’s as if the loss value no longer serves as a reliable indicator of whether a sample belongs to the majority or minority groups. In other words, when the spurious correlation is low, a sample’s misclassification could be due to factors other than the model’s reliance on the spurious features of a minority sample.

In low spurious correlations like 70\%, AFR has the best WGA. As we have stated before, it is because of the way we evaluate AFR, and give it an advantage
AFR probably underperforms in high spurious correlation settings because they do not have any consideration for the number of examples from the majority group. So, even if the underweight samples with lower loss are probably from the majority group, the overall effect of the majority samples still remains high. On the other hand, as LFR selects an equal number of samples from both the majority and minority, it does not have the aforementioned problem.

\begin{figure}
\begin{center}
\fbox{
\includegraphics[width=0.9\linewidth]{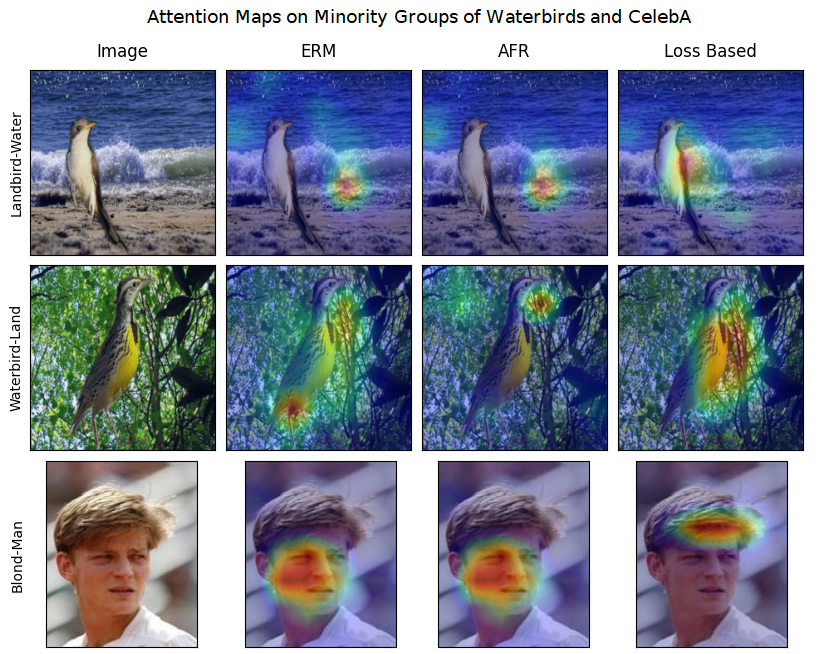}
}
\end{center}
   \caption{Saliency map of models before (ERM) and after last layer retraining with AFR and LFR. LFR has a better quality in terms of pushing the model to attend to core attributes.}
\label{fig:saliency}
\end{figure}

We also visualize our method's attention map with Grad-Cam~\cite{jacobgilpytorchcam} to compare it with other models. As it can be seen in figure \ref{fig:saliency},  using LFR, the model attends more to core attributes, in contrast to the pre-trained ERM model and AFR. More results are available in \ref{saliency}.

\section{Conclusion}
We proposed loss-based feature re-weighting, a method for selecting a few images from a small proportion of training data to retrain the last linear of the model which improves robustness to spurious correlation. While LFR is fast and straightforward to use, it achieved SOTA in many various datasets in comparison to previously proposed methods and baselines, even works with the assumption of group annotations availability, in addition to better performance in attending core attributes in input images.

{\small
\bibliographystyle{ieee_fullname}
\bibliography{egbib}
}
\clearpage
\appendix

\section{Datasets}\label{dataset_det}

The table \ref{table:datasets} demonstrates the number of samples in each group for \textit{Tr} and \textit{LL} partitions.

\begin{table*}[ht]
\caption{Number of samples in each dataset in the experiments.}
\centering
\begin{tabular}{lccccccccl} 
 \toprule
 Dataset & Split & Label & Spurious & 70\% & 80\% & 90\% & 95\% & 99\% & 100\%\\ [0.5ex] 
 \midrule
 \multirow{8}{*}{Waterbirds}
     & \multirow{4}{*}{Train} 
         & \multirow{2}{*}{Landbird} & Land & 872 & 881 & 857 & 865 & 864 & 865 \\ \cmidrule{4-10}
         &&&  Water & 363 & 200 & 104 & 45 & 9 & 0  \\
         \cmidrule{3-10}
         && \multirow{2}{*}{Waterbird} &Land & 370 & 216 & 96 & 45 & 8 & 0 \\ \cmidrule{4-10}
         &&& Water & 865 & 865 & 865 & 865 & 865 & 865 \\
     \cmidrule{2-10}
     &\multirow{4}{*}{Last Layer}
         & \multirow{2}{*}{Landbird} & Land & 202 & 192 & 194 & 723 & 723 & 723 \\ \cmidrule{4-10}
         &&& Water & 72 & 48 & 19 & 33 & 33 & 33  \\
         \cmidrule{3-10}
         && \multirow{2}{*}{Waterbird} & Land & 82 & 48 & 21 & 11 & 11 & 11 \\ \cmidrule{4-10}
         &&& Water & 192 & 192 & 192 & 192 & 192 & 192 \\
         \midrule
         
 \multirow{8}{*}{CelebA}
     & \multirow{4}{*}{Train} 
         & \multirow{2}{*}{Black Hair} & Female & 13524 & 11762 & 10567 & 10018 & 9636 & 9346 \\
         \cmidrule{4-10}
         &&& Male & 12561 & 11063 & 9721 & 9203 & 8808 & 8914  \\
         \cmidrule{3-10}
         && \multirow{2}{*}{Blond Hair} & Female & 18260 & 18260 & 18260 & 18260 & 18260 & 18260 \\
         &&& Male & 7825 & 4565 & 2028 & 961 & 184 & 0 \\
     \cmidrule{2-10}
     &\multirow{4}{*}{Last Layer}
         & \multirow{2}{*}{Black Hair} & Female & 3386 & 3003 & 2632 & 2477 & 2498 & 2520 \\
         \cmidrule{4-10}
         &&& Male & 3214 & 2772 & 2501 & 2386 & 2365 & 2343  \\
         \cmidrule{3-10}
         && \multirow{2}{*}{Blond Hair} & Female & 4620 & 4620 & 4620 & 4620 & 4620 & 4620 \\\cmidrule{4-10}
         &&& Male & 1980 & 1155 & 513 & 243 & 243 & 243 \\
     \bottomrule
\end{tabular}
\label{table:datasets}
\end{table*}

\section{Setup}\label{setup_det}
To fine-tune our method's hyperparameters, we carefully considered several important factors. Specifically, when adjusting the sample size (the number of samples chosen from misclassified and correctly-classified groups for each label), our aim was to ensure that the final dataset exhibited a rough balance among different group labels. For example, when working with the Waterbirds dataset, we found that using a sample size of about 30 was sufficient to achieve state-of-the-art performance after re-training. This meant we selected 120 data points using our loss-based method. Additionally, we incorporated a weight decay of 1e-3 or 1e-4 and experimented with learning rates ranging from 1e-4 to 1e-3 to further enhance our model's performance. On the other hand, for the CelebA dataset, which had more data points, we needed a sample size of nearly 700 to get good results.

\begin{figure}[hbt]
\begin{center}
\includegraphics[height=\linewidth]{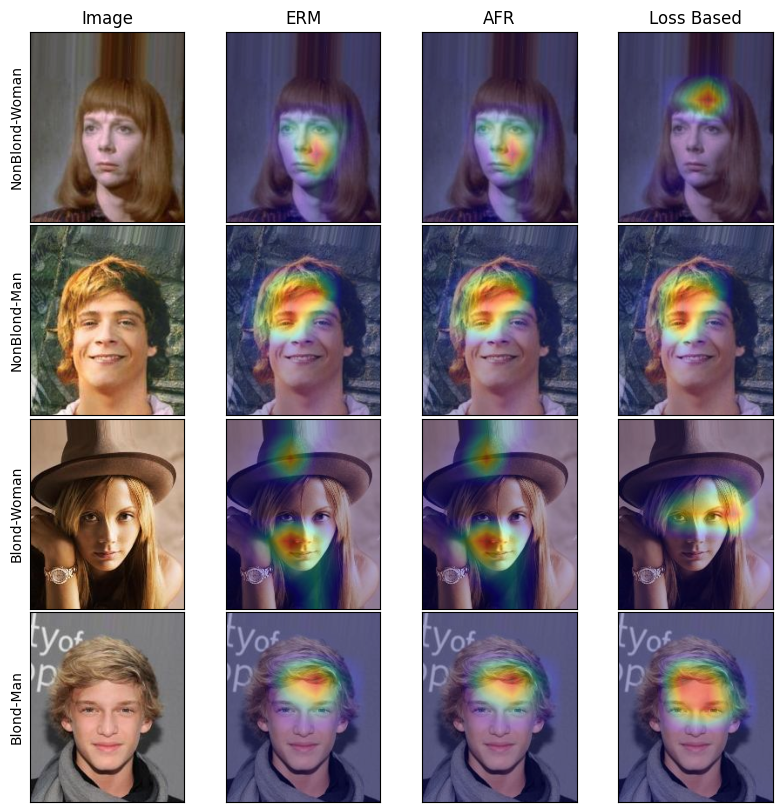}
\end{center}
   \caption{Additional GradCAM visualizations for CelebA dataset. Both ERM and AFR models output higher salience for non-causal attributes. LFR corrects this both for majority and minority examples.}
\label{fig:celebamap}
\end{figure}

\begin{figure}[hbt]
\begin{center}
\includegraphics[height=\linewidth]{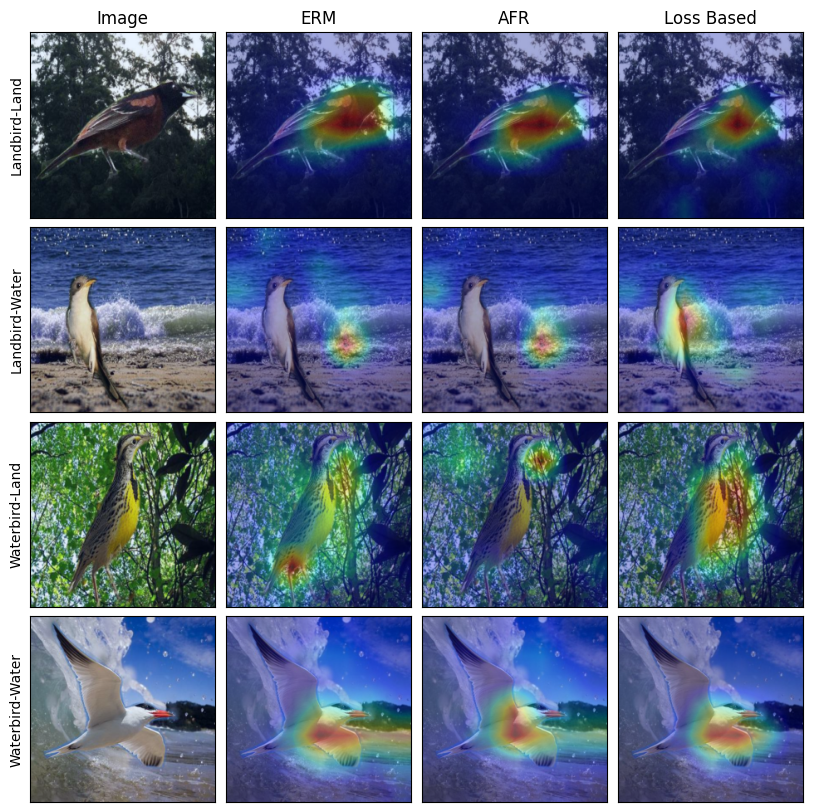}
\end{center}
   \caption{Additional GradCAM visualizations for Waterbirds dataset. Both ERM and AFR models output higher salience for spurious attributes (here is the background). LFR corrects this both for majority and minority examples.}
\label{fig:waterbirdsmap}
\end{figure}

\section{Saliency Maps of Model's attention regions}\label{saliency}

We have incorporated additional visualizations to support the effectiveness of our method. By employing Grad-CAM~\cite{Selvaraju_2019}, we illustrate that models trained with ERM tend to classify inputs based on spurious features. Additionally, AFR appears to rely on non-causal features for prediction, possibly due to the regularization term in its loss function, which encourages the weights to stay close to those of ERM. LFR rectifies this phenomenon and relies on core features for both majority and minority group examples. Figure \ref{fig:celebamap} and \ref{fig:waterbirdsmap} provide visual representations of these trends on the CelebA and Waterbirds datasets, respectively, vividly demonstrating how our approach effectively addresses these issues.

\end{document}